\documentclass[letterpaper, 10pt, conference]{ieeeconf}  % Comment this line out if you need a4paper

\IEEEoverridecommandlockouts                              % This command is only needed if 
\usepackage{graphicx} % for pdf, bitmapped graphics files
\usepackage{amsmath} % assumes amsmath package installed
\usepackage{amssymb}  % assumes amsmath package installed

\usepackage{hyperref}
\usepackage{cleveref}

\newcommand\Tstrut{\rule{0pt}{2.6ex}}  
\usepackage{pifont}% http://ctan.org/pkg/pifont
\newcommand{\norm}[1]{\left\lVert#1\right\rVert}

\usepackage{cite}

\title{\LARGE \bf
EgoNN: Egocentric Neural Network for Point Cloud Based 6DoF Relocalization at the City Scale*
}
\author{Jacek Komorowski$^{1}$, Monika Wysoczanska$^{1}$ and Tomasz Trzcinski$^{1}$% <-this % stops a space
\thanks{*The project was funded by POB Research Centre for Artificial Intelligence and Robotics of Warsaw University of Technology within the Excellence Initiative Program - Research University (ID-UB)}% <-this % stops a space
\thanks{$^{1}$Authors are with the Faculty of Electronics and Information Technology,
Warsaw University of Technology, Poland
{\tt\small \{firstname.lastname\}@pw.edu.pl}}%
}

\begin{document}

\maketitle
\thispagestyle{empty}
\pagestyle{empty}

%%%%%%%%%%%%%%%%%%%%%%%%%%%%%%%%%%%%%%%%%%%%%%%%%%%%%%%%%%%%%%%%%%%%%%%%%%%%%%%%
\begin{abstract}

The paper presents a deep neural network-based method for global and local descriptors extraction from a point cloud acquired by a rotating 3D LiDAR.
The descriptors can be used for two-stage 6DoF relocalization. First, a course position is retrieved by finding candidates with the closest global descriptor in the database of geo-tagged point clouds. Then, the 6DoF pose between a query point cloud and a database point cloud is estimated by matching local descriptors and using a robust estimator such as RANSAC.
Our method has a simple, fully convolutional architecture based on a sparse voxelized representation. 
It can efficiently 
extract a global descriptor and a set of keypoints with local descriptors from large point clouds with tens of thousand points.
Our code and pretrained models are publicly available on the project website.~\footnote{\url{https://github.com/jac99/Egonn}}

\end{abstract}

%%%%%%%%%%%%%%%%%%%%%%%%%%%%%%%%%%%%%%%%%%%%%%%%%%%%%%%%%%%%%%%%%%%%%%%%%%%%%%%%

\section{Introduction}

Relocalization at a city-scale is an emerging task with various applications in robotics and autonomous vehicles, such as loop closure in SLAM or \textit{kidnapped robot} problem~\cite{chen2020overlapnet}.
%, when no prior information about the position in a map is given 
A typical approach is a two-step process: 1) coarse localization using global descriptors, 2) precise 6DoF pose estimation by pairwise registration.

The development of LiDAR technologies enabled the shift from an image-based approaches, affected by scene appearance changes, to 3D methods relying on the scene geometry.
Several learning-based methods for LiDAR-based global descriptor extraction were recently proposed~\cite{uy2018pointnetvlad, du2020dh3d, Komorowski_2021_WACV}, but they are intended for relatively small point clouds with few thousand points. 
In contrast, scans from modern 3D LiDARs, such as Velodyne HDL-64E, show 360$^{\circ}$ view of the scene and contain tens of thousand points. For computational efficiency, methods operating on such large point clouds typically convert them to some form of intermediate representation, such as a set of multi-layer 2D images~\cite{xu2021disco}, before further processing. 
This hides some information about a scene structure and can adversely impact the performance.

In this work, we address the problem of a point cloud-based relocalization at a city scale. 
We consider a typical scenario~\cite{du2020dh3d}, where the map consists of LiDAR scans with known 6DoF absolute pose, and the aim is to find a 6DoF pose of a query scan.
We propose a network architecture to efficiently extract both a global descriptor for coarse-level place recognition and a set of keypoints with their local descriptors for 6DoF pose estimation. 
The network is designed to efficiently process large point clouds, with tens of thousand points, acquired by modern rotating LiDAR sensors.
A similar approach is proposed in~\cite{du2020dh3d}, but their DH3D method handles smaller point clouds with less than 10k points.
The design of our method is inspired by the state-of-the-art MinkLoc3D~\cite{Komorowski_2021_WACV} global point cloud descriptor. It processes raw 3D point clouds using a sparse voxelized representation and 3D convolutional architecture without the need for a point cloud conversion to an intermediate form. 
However, MinkLoc3D is intended for smaller point clouds with few thousand points.
In this work, we investigate the scalability of this architecture to much larger point clouds.

The main contribution of our work is the development of an efficient network architecture for extraction of both a global descriptor for coarse-level place recognition and a set of keypoints with their local descriptors for a precise 6DoF pose estimation. The method efficiently  processes raw point clouds acquired by modern 360$^{\circ}$ rotating LiDAR sensors, with tens of thousand points.
Additionally, we make our code, pre-trained models, and training/evaluation splits publicly available. We believe this will help advance state of the art by allowing the community to evaluate future methods against the same baselines.

\section{Related work}

Early methods to generate a global point cloud descriptor use handcrafted features.
%Among the proposed methods for global place recognition based on 3D point clouds, handcrafted descriptors constitute a significantly large group up to date. 
Most of the approaches~\cite{knopp2010hough, tombari2011combined, steder2011place, he2016m2dp} use statistical calculations and histogram aggregations to describe a scene. ScanContext~\cite{kim2018scan} represents a scene as a bird's-eye-view RGB image that  describes geometrical information of an egocentric environment. Seed~\cite{fan2020seed} extends this idea by first segmenting a point cloud, and Weighted ScanContext~\cite{cai2021weighted} uses the intensity information of points to enhance geometric features. 
LiDAR Iris~\cite{wang2019lidar} introduces a binary signature image computed from a bird's-eye-view point cloud representation. 
% Recently, \cite{shan2021robust} proposes a bag-of-words based approach on extracted ORB feature descriptors from the intensity image projected from a 3D point cloud.

The alternative group of methods leverages deep neural networks to produce a discriminative global descriptor in a learned manner. 
Most of them operates on relatively small point clouds built by accumulating consecutive 2D LiDAR scans during the vehicle traversal and downsampling the merged cloud. 
For instance, PointNetVLAD~\cite{uy2018pointnetvlad}, one of the first learned global descriptors, operates on point clouds spanning 20\,m in the longitudinal direction and downsampled to 4096 points. 
Several methods~\cite{liu2019lpd, du2020dh3d, Komorowski_2021_WACV} improve upon NetVLAD, but all of them are operate on point clouds with the same characteristic. They poorly scale to larger clouds, with tens of thousands points and covering the area with over 100\,m radius, acquired using modern 3D rotating LiDARs.
% \todo[inline]{Give some evidence for this claim; maybe there's deteriorated performance of some of these methods on Kitti? Maybe we van run eval of MinkLoc3D in experimental evaluation section? Also discuss why this happens - maybe they are too big to efficiently process such large point clouds? And architectures have limited capability to extract informative features from large scale point clouds. Possibly we can discuss architectures of these methods here.}

For computational efficiency, methods operating on larger scans usually convert them to an intermediary representation before processing with a deep neural network. 
OREOS~\cite{8968094} projects the input into a 2D range image using a spherical projection model. 
OverlapNet~\cite{chen2020overlapnet} extracts four different 2D representations: range, intensity values, normals, and semantic classes. 
DiSCO~\cite{xu2021disco} converts a point cloud into a set of multi-layer 2D images in cylindrical coordinates.
Such conversion loses some information about the scene structure, which may adversely impact the performance.
Our method operates on raw point clouds from a modern 3D LiDAR, without the need for an intermediary representation. 

In addition to a global descriptor extraction, some recent methods produce additional information that can be used for an initial alignment of two point clouds before the final registration using an ICP~\cite{besl1992method} or similar method.
OREOS~\cite{8968094} and DiSCO~\cite{xu2021disco} estimate a relative yaw angle between two scans.
In addition to yaw angle, OverlapNet~\cite{chen2020overlapnet} estimates an overlap between two point clouds.
Our method regresses a set of keypoints with their descriptors that can be used to directly estimate relative 6DoF pose without the need for an initial alignment.

The closest approach to ours is DH3D~\cite{du2020dh3d}, which is the first method that unifies global descriptor extraction with local features computation. 
However, DH3D is handles relatively small point clouds with 8k points and spanning a 20\,m distance in the longitudinal direction. Our method scales to much larger clouds with tens of thousand points.
%with no additional preprocessing.

Some recent deep learning-based point cloud registration methods~\cite{wang2019deep,yew2020rpm} directly estimate relative 6DoF pose between two point clouds without finding correspondences between local features.
However, results in~\cite{9320424} suggest that direct keypoint correspondence estimation can more accurately register two partially overlapping point clouds and better handle larger initial misalignment.
Hence our method computes repetitive keypoints and discriminative local descriptors that can be used for a robust RANSAC-based 6DoF pose estimation.

\section{EgoNN: Egocentric neural network for global and local descriptors}

We propose a neural network architecture, dubbed \emph{EgoNN}, to compute a global descriptor and a set of keypoints with local descriptors from an input point cloud.
Descriptors and keypoints are extracted in a single pass through the network, with a majority of computations shared between the global and local parts.
The method can be used for efficient two-step 6DoF relocalization: coarse localization with a global descriptor and 6DoF pose estimation using local features.
%The relocalization approach is illustrated in Fig.
%~\todo[inline]{Consider visualization of a relocalization approach and adding a figure similar to the one in DH3D}.

The input to our method is a 3D point cloud acquired by a rotating 3D LiDAR sensor.
%The LiDAR scan is defined as a set of 3D points $\mathcal{P} = \left\{ \mathrm{p}_1, \mathrm{p}_2, \ldots, \mathrm{p}_N \right\}$, where $N$ is the number of points.
We apply a common preprocessing to remove points on the ground plane level and below, based on the z coordinate. 
These points are non-informative for localization purposes, and their removal allows more efficient processing.

We convert the input point cloud to cylindrical coordinates to improve rotational invariance, necessary for loop closure and place recognition applications.
A point $(x, y, z)$ in Cartesian coordinates is converted to $(\rho, \theta, z)$ in cylindrical coordinates, where
$\rho = \sqrt{x^2 + y^2}$ and $\theta = \arctan \left( y / x \right)$.

\subsection{Network Architecture}

EgoNN operates on a sparse voxelized representation of the input point cloud.
The input point cloud, in cylindrical coordinates, is quantized into a single channel sparse tensor. 
The values of this single channel are set to one for occupied voxels. 
In our implementation we use: $\rho_s=0.3$\,m, $\theta_s=1^\circ$ and $z_s = 0.2$\,m quantization steps for angular, radial (in x-y plane) and z coordinates respectively.
The network produces a global descriptor of the input point cloud; a set of regressed keypoints, 
their saliency uncertainty estimates and descriptors.

\begin{figure}[htb!]
\centering
\includegraphics[width=.95\linewidth,trim={0 1.2cm 0 0},clip]{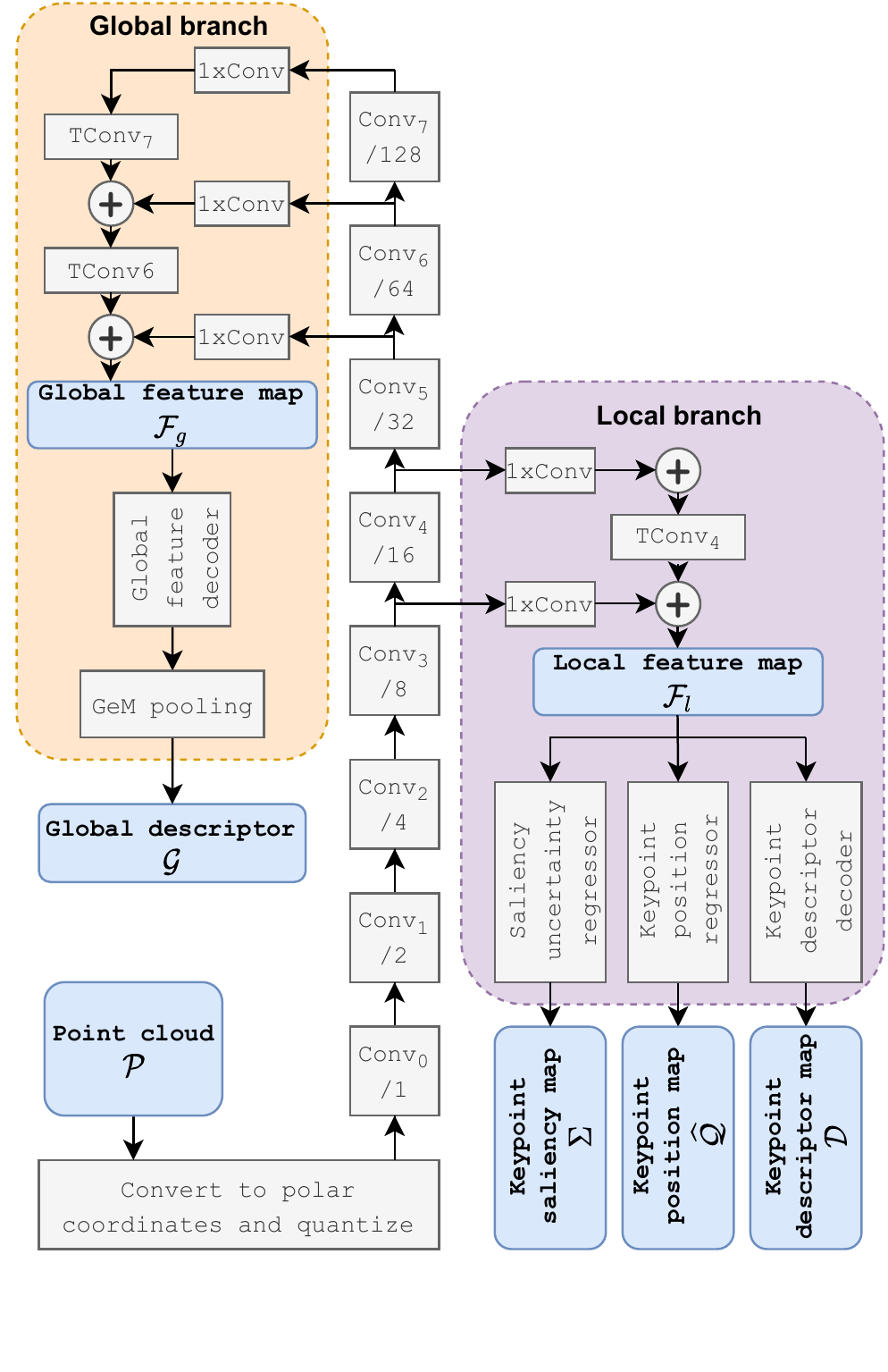}
\caption{High-level architecture of EgoNN network.
Numbers with a slash (e.g. /32) denote a stride of a feature map produced by each convolutional block in the network trunk.
}
\label{fig:high_level}
\end{figure}

% NOW DESCRIBE DETAILS OF THE NETWORK
The network has a 3D convolutional architecture shown in Fig.~\ref{fig:high_level}.
The design is inspired by a successful MinkLoc3D~\cite{Komorowski_2021_WACV} global point cloud descriptor.
To improve the performance on larger point clouds, we increased the depth of the network and added channel attention (ECA~\cite{wang2020eca}) to convolutional blocks.
%It's a 3D convolutional network modelled after Feature Pyramid Network~\cite{lin2017feature} architecture.
The bottom-up trunk consists of eight convolutional blocks 
%$\mathcal{C}_0, \ldots, \mathcal{C}_7$ 
$\mathrm{Conv}_0, \ldots, \mathrm{Conv}_7$ 
producing sparse 3D feature maps with decreasing spatial resolution and increasing receptive field.
Each convolutional block, starting from $\mathrm{Conv}_1$, decreases the spatial resolution by two. %Numbers with slash (e.g. /32) in Fig.~\ref{fig:high_level} denote a stride of a feature map produced by each block.
There are two top-down parts: global and local branch, both made of transposed 3D convolutions.
Feature maps from higher pyramid levels, upsampled using transposed convolution (denoted $\mathrm{TConv}$), are added to the skipped features from the corresponding layer in the bottom-up trunk using lateral connections. 
Lateral connections apply convolutions with 1x1x1 kernel (denoted $\mathrm{1xConv}$) to unify the number of channels produced by bottom-up blocks before they are merged in the top-down pass.
Such design produces feature maps with relatively high spatial resolution and a large receptive field, capturing high-level semantics of the input point cloud. 
%Experimentally it proved advantageous over a simple convolutional architecture without top-down processing. 
%\todo[inline]{Add relevant experiment to ablation study}

% GLOBAL branch
\textbf{Global branch} computes a global point cloud descriptor.
It merges feature maps from higher pyramid levels to produce a \emph{global feature map} $\mathcal{F}_g \in \mathbb{R}^{K \times 128}$.
%with $K$ 128-dimensional features.
%$K$ is dependant on the size and distribution of points in the input point cloud.
Each $\mathcal{F}_g$ element is processed by a \emph{global feature decoder}, a two-layer MLP.
The resulting 256-dimensional feature map is pooled with generalized-mean (GeM)~\cite{radenovic2018fine} pooling to produce a global point cloud descriptor $\mathcal{G} \in \mathbb{R}^{256}$.

% LOCAL DESCRIPTOR BRANCH
\textbf{Local branch} computes keypoints, their saliency uncertainty estimates, and descriptors. 
We adapt the approach proposed in USIP~\cite{li2019usip} for keypoint detection, which fits well with our architecture. 
%USIP requires dividing a point cloud into a regular grid of cells, and regresses one keypoint per each cell. 
%Such division is obtained when processing a point cloud with 3D convolutional architecture without extra cost. 
%Spatial locations in a higher level feature maps produced by convolutional blocks in the bottom-up trunk correspond to disjoint regions of the input point cloud of the size proportional to the pyramid level.
The local branch merges feature maps from lower pyramid levels (level 3 and 4) to produce a \emph{local feature map} $\mathcal{F}_l \in \mathbb{R}^{M \times 64}$.
Each $\mathcal{F}_l$ element is processed by three heads.
\emph{Keypoint position regressor}, a two layer MLP followed by tanh function, computes keypoint coordinates map $\widehat{\mathcal{Q}} \in \left( -1, 1 \right)^{M \times 3}$.
\emph{Saliency uncertainty regressor}, a two layer MLP followed by a softplus, produces a keypoint uncertainty map $\Sigma \in \mathbb{R}_{+}^{M}$.
\emph{Keypoint descriptor decoder}, a two layer MLP, yields a keypoint descriptor map  $\mathcal{D} \in \mathbb{R}^{M \times 128}$.
%, with 128-dimensional local descriptors of regressed keypoints.
$M$ is the number of \emph{supervoxels} in the local feature map $\mathcal{F}_l$, that is the number of non-empty spatial locations in $\mathcal{F}_l$. In our implementation, supervoxel covers 8x8x8 voxels.

\begin{figure}[tbh]
\centering
\includegraphics[width=.7\linewidth,trim={0.5cm 0.2cm 0.5cm 0.3cm},clip]{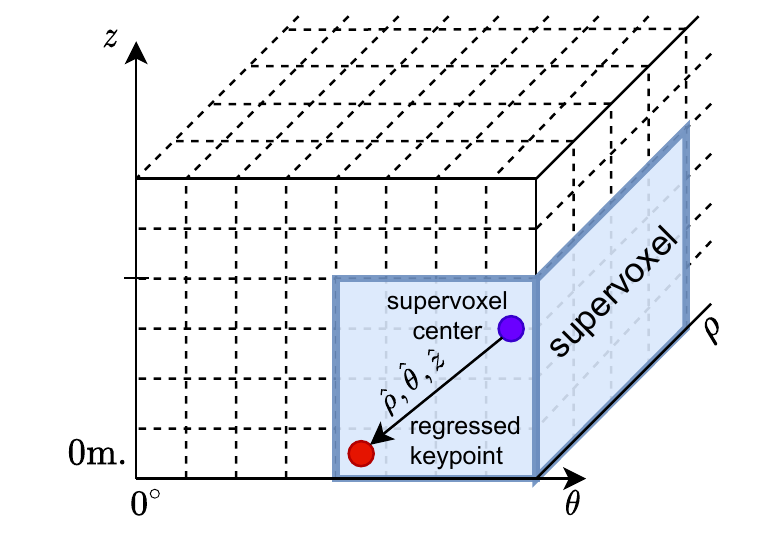}
\caption{Illustration of the idea behind a keypoint position regressor. Cylindrical coordinates $\left( \hat{\rho}, \hat{\theta}, \hat{z} \right)$ of one keypoint relative to the supervoxel center are regressed in each non-empty supervoxel. Dashed lines indicate voxel boundaries.}
\label{jk:fig:supervoxel}
\end{figure}

Keypoint position regressor outputs normalized keypoint cylindrical coordinates $\left( \hat{\rho}, \hat{\theta}, \hat{z} \right) \in \widehat{\mathcal{Q}}$ relative to the supervoxel center. One keypoint is regressed in each non-empty supervoxel.
See Fig.~\ref{jk:fig:supervoxel} for illustration.
To compute Cartesian keypoint coordinates $\mathcal{Q}$ we first transform them into absolute cylindrical coordinates $\left( \rho, \theta, z \right)$:
\begin{equation}
    \label{jk:eq:rel_cylindrical_to_abs}
    \rho =  \frac{\hat{\rho} s_\rho}{2}  + \rho_c \, ,
    \theta =  \frac{\hat{\theta} s_\theta}{2} + \theta_c \, ,
    z = \frac{\hat{z} s_z}{2} + z_c \, ,
\end{equation}
where $s_\rho, s_\theta, s_z$ is the supervoxel size in each dimension 
and $\left( \rho_{c}, \theta_{c}, z_{c} \right)$ are absolute cylindrical coordinates of the supervoxel center.
In our implementation local feature map $\mathcal{F}_l$ has a stride 8.  So the supervoxel size in each direction is eight times the initial quantization step, that is $s_\rho=2.4 \mathrm{m}, s_\theta=8^{\circ}, s_z=1.6 \mathrm{m}$.
Then, we transform cylindrical coordinates into Cartesian: $x = \rho \cos \theta$, $y = \rho \sin \theta$.

Details of each network block are shown in Table~\ref{jk:tab-details}. 
$\mathrm{Conv}_k$ and $\mathrm{1Conv}$ blocks contain 3D convolutional layers; $\mathrm{TConv}_k$ are 3D transposed convolutions;
MLP denotes a multi-layer perceptron with the number of neurons in each layer listed in parentheses.
%, BN is a BatchNorm and ECA is Efficient Channel Attention~\cite{wang2020eca}.

\begin{table}
	\centering
	\caption{Details of EgoNN architecture. BN=BatchNorm, MLP=Multi-Layer Perceptron.}
    \begin{tabular}{@{\enskip}l@{\enskip}l@{\enskip}}
     \hline
    Block  &  Layers\\
     \hline
    \multicolumn{2}{c}{\textbf{Bottom-up trunk}} \\
    {\tt Conv$_0$}  &  32 filters 5x5x5 - BN - ReLU\\
%    {\tt conv$_k$}, $k = 1, \ldots, 7$  &  32 filters 2x2x2 stride 2   \\
    %\Tstrut
    {\tt Conv$_k$} &  $c_k$ filters 2x2x2 stride 2  - BN - ReLU \\
           &  $c_k$ filters 3x3x3 stride 1 - BN - ReLU  \\
           &  $c_k$ filters 3x3x3 stride 1 - BN - ReLU\\
           & ECA (Efficient Channel Attention) \\
        \multicolumn{2}{c}{where $c_1=32, c_2 = c_3=64, c_4 = c_5 = c_6 = c_7=128$} \\
    \hline
    \multicolumn{2}{c}{\textbf{Global branch}} \\
    {\tt TConv$_k$}, $k=6,7$  &  128 filters 2x2x2 stride 2  \\
    %\Tstrut
    {\tt 1xConv} &  128 filters 1x1x1 stride 1  \\
    %\Tstrut
    {\tt Global feature} & MLP(192, 256) \\
    {\tt decoder} & \\
    %\Tstrut
    {\tt GeM pooling} & Generalized-mean Pooling~\cite{radenovic2018fine} layer \\
    \hline
    \multicolumn{2}{c}{\textbf{Local branch}} \\
    {\tt TConv$_k$}, $k=4,5$  &  64 filters 2x2x2 stride 2  \\
    %\Tstrut
    {\tt 1xConv}  &  64 filters 1x1x1 stride 1  \\
    %\Tstrut
    {\tt Saliency uncertainty} & MLP(32, 1) - Softplus \\
    {\tt regressor} \\
    %\Tstrut
    {\tt Keypoint position} & MLP(32, 3) - Tanh\\
    {\tt regressor} \\
    %\Tstrut
    {\tt Keypoint descriptor} & MLP(96, 128) -  $L_2$-normalization \\
    {\tt decoder} \\
    \hline
	\end{tabular}
    \label{jk:tab-details}
\end{table}

\subsection{Network Training}
%The entire network is trained in an end-to-end manner.
To reduce requirements for computational resources, each training step is split into two substeps. 
First, one mini-batch of training examples is used to calculate the loss on the global descriptor, and network weights are updated. Then, a different mini-batch is used to compute the loss on local features, and the network weights are updated again.
Global part is trained using relatively large batches, allowing online mining of hard and informative training triplets.
%and improves the global descriptor performance.
Local part training requires more GPU resources; thus, smaller batch size is used.
Note that weights of the network trunk are optimized in both phases, as the gradient of both global and local losses backpropagates through them.

\textbf{Global branch training substep.}
Mini-batches are constructed by sampling 128 pairs of similar point clouds. 
Two point clouds are considered similar if the distance between their centers, computed using the  ground truth poses, is at most 2\,m. 

Mini-batch is processed by the network to compute a global descriptor $\mathcal{G}$ of each batch element, while the local branch is disabled.
Mini-batch elements are arranged into triplets.
For $i$-th batch element, we construct a training triplet $(i, p_i, n_i)$,
where $i$ is an index of an anchor element, $p_i$ is an index of a point cloud similar to the anchor 
%(by construction there's at least one such point cloud in the batch) 
and $n_i$ is an index of a point cloud dissimilar to the anchor.
Point clouds are dissimilar if the distance between their centers is above 10\,m.
We use the batch-hard mining~\cite{hermans2017defense} strategy to construct informative triplets and choose the hardest positive and negative examples within a batch.
We augment training point clouds by removing points within a randomly selected cuboid, jittering point positions by adding Gaussian noise with $\sigma=0.1$, and applying random rotation around the z-axis.
%This improves rotational invariance of the resultant global descriptor and reduces the overfitting.

%To compute a discriminative global descriptor we use a triplet margin loss~\cite{hermans2017defense} defined as:
We use a triplet margin loss~\cite{hermans2017defense} defined as:
\begin{equation}
\mathcal{L}_{G}(a_i,p_i,n_i) = \max \left\{ \norm{a_i - p_i}_2 - \norm{a_i, n_i}_2 + m, 0 \right\} ,
\label{eq:loss_global}
\end{equation}
where 
\(a_i, p_i, n_i\) are embeddings of an anchor, a positive and negative elements in $i$-th training triplet and $m=0.2$ is a margin hyperparameter. 

%\todo[inline]{Possibly mention dynamic batch sizing}
%Triplets $(i, p_i, n_i)$ are used to compute a global descriptor loss $\mathcal{L}_{\mathrm{G}}$.
%Pairs $(i, p_i)$ are used to compute a local descriptor %$\mathcal{L}_{\mathrm{D}}$ and keypoint loss $\mathcal{L}_{\mathrm{K}}$.

\textbf{Local branch training substep.} In each training step, we sample a pair of similar point clouds, that is point clouds with at most 2\,m distance between their centers.
We augment training point clouds by random rotation around the z-axis and random translation in the x-y plane up to 5\,m. %This improves local descriptors rotational invariance and reduces the overfitting.

Sampled mini-batch is processed by the network to compute keypoint descriptor map $\mathcal{D}$, keypoint position map $\mathcal{Q}$ and saliency uncertainty map $\Sigma$ for each batch element.
The global branch is switched off.

The local loss function $\mathcal{L}_L$ is a sum of three terms: 
two keypoint-related losses $\mathcal{L}_C$, $\mathcal{L}_{P2P}$ and a local descriptor loss $\mathcal{L}_D$:

\begin{equation}
\mathcal{L} = \lambda_C \mathcal{L}_C + \lambda_{P2P} \mathcal{L}_{P2P} + \lambda_D \mathcal{L}_D,
\label{eq:loss_total}
\end{equation}
where $\lambda_C=1, \lambda_{P2P}=1, \lambda_D=1 \in \mathbb{R}_{+}$ are experimentally chosen hyperparameters.
 
\textbf{Keypoint-related loss.}
To compute the keypoints position, we adapt the approach proposed in USIP~\cite{li2019usip}.
One keypoint is regressed in each non-empty supervoxel (see Fig.~\ref{jk:fig:supervoxel}).
To get repeatable keypoints, each keypoint regressed in the first point cloud (in the area overlapping with the second point cloud) should have a corresponding keypoint detected in the second cloud. 
Additionally, a position of the regressed keypoint should be close to some 3D point in the input cloud.
To enforce these constraints, the keypoint loss is a sum of two terms: 
$\mathcal{L}_C$
is the probabilistic chamfer loss~\cite{li2019usip} that minimizes the probabilistic
distances between corresponding pairs of keypoints in two point clouds.
It drives corresponding keypoints to be as close as possible, taking into account their saliency uncertainty. 
$\mathcal{L}_{P2P}$ is the point-to-point loss that minimizes the distance between regressed keypoints and their nearest neighbors in the point cloud. 
%It drives each estimated keypoint to be close to a point in the input cloud. 
See USIP~\cite{li2019usip} for derivation and discussion on keypoint-related loss terms.
Let $\mathcal{Q}=\left\{ \mathrm{q}_1, \ldots,  \mathrm{q}_N \right\}$ and 
%\mathrm{q}_i \in \mathbb{R}^3
$\widetilde{\mathcal{Q}}=\left\{ \widetilde{\mathrm{q}}_1, \ldots, \widetilde{\mathrm{q}}_M \right\}$,
%\widetilde{\mathrm{q}}_i \in \mathbb{R}^3$
be sets of keypoint Cartesian coordinates computed for a pair of overlapping point clouds.
$\Sigma=\left\{ \sigma_1, \ldots, \sigma_N \right\}, \sigma_i \in \mathbb{R}_{+}$ and 
$\widetilde{\Sigma}=\left\{ \widetilde{\sigma}_1, \ldots, \widetilde{\sigma}_M \right\}_{i=1}^{M}, \widetilde{\sigma}_i \in \mathbb{R}_{+}$ their saliency uncertainty estimates.
$\mathcal{Q}'=\left\{ \mathrm{q}'_1, \ldots, \mathrm{q}'_M \right\}$ is the set of keypoints in the second point cloud $\widetilde{\mathcal{Q}}$ brought to the first point cloud coordinate frame, using the ground truth transform $\mathrm{T}$, $\mathcal{Q}' = \mathrm{T}^{-1} \circ \widetilde{\mathcal{Q}}$.

%That is, the distance between a keypoint in the first point cloud, to the closest keypoint detected in the second point cloud should be as small as possible. And vice verse.
%The intuition behind the probabilistic chamfer loss is that it drives keypoints regressed from two point clouds to be located at the same places.

Probabilistic chamfer loss~\cite{li2019usip} $\mathcal{L}_{\mathrm{c}}$ is defined as:
\begin{equation}
\mathcal{L}_C =
\sum_{i=1}^N \left( 
\ln s_{i} + \frac{d_{i}}{s_{i}}
\right)
+
\sum_{j=1}^M \left( 
\ln s'_{j} + \frac{d'_{j}}{s'_{j}}
\right) ,
\label{eq:loss_chamfer}
\end{equation}
where 
$d_{i} = \norm{\mathrm{q}_i - \mathrm{q}'_{nn(i)}}_2$ and
$d'_{j} = \norm{\mathrm{q}'_j - \mathrm{q}_{nn'(j)}}_{2}$ 
are Euclidean distances between corresponding keypoints detected in two point clouds;
%keypoints in one point cloud and their closests keypoints in the other cloud;
%is the distance between $j$-th keypoint in the second cloud and the closest keypoint in the first cloud,
%is the Euclidean distance between $i$-th keypoint detected in the first point cloud and the closest keypoint in the second point cloud,
$s_i = \frac{1}{2} \left( \sigma_i + \widetilde{\sigma}_{nn(i)} \right)$,
$s'_j = \frac{1}{2} \left( \widetilde{\sigma_j} + \sigma_{nn'(j)} \right)$
are mean saliency uncertainties of corresponding keypoints;
$nn(i) = \arg \min_j \norm{\mathrm{q}_i - \mathrm{q}'_j}_2$ is the index of the keypoint in the second point cloud corresponding to the $i$-th keypoint in the first point cloud
and
$nn'(j) = \arg \min_i \norm{\mathrm{q}'_j - \mathrm{q}_i}_2$ is the index of the keypoint in the first point cloud corresponding to the $j$-th keypoint in the second point cloud.

Point-to-point loss $\mathcal{L}_{\mathrm{p2p}}$ is defined as:
\begin{equation}
\mathcal{L}_{P2P} =
\sum_{i=1}^N \min_{\mathrm{p}_j \in \mathcal{P}}
\norm{\mathrm{q}_i - \mathrm{p}_j}_2
+
\sum_{j=1}^M \min_{\widetilde{\mathrm{p}}_i \in \widetilde{\mathcal{P}}}
\norm{\widetilde{\mathrm{q}}_j - \widetilde{\mathrm{p}_i}}_2 ,
\label{eq:loss_p2p}
\end{equation}
where each of two terms is a sum of distances between regressed keypoints and their nearest neighbours in the input point cloud.

\textbf{Local descriptors-related loss.}
%\textbf{Local descriptor loss.}
%Hence we adopt the approach of directly learning discriminative local descriptors for establishing keypoints correspondence. 
We compute local descriptors not for all points in the input cloud but for a much smaller, typically by an order of magnitude, set of regressed keypoints. This allows efficient processing of relatively large points clouds.
Similarly as in~\cite{9320424}, keypoint correspondence is posed as a multi-class classification problem, where a point in the first cloud is classified as corresponding to one point in the second cloud.
%\todo[inline]{Discuss somewhere in the paper, what's the typical ratio of all points (before and after the quantization) to the number of all regressed keypoints and to the number of keypoints selected for RANSAC estimation.}
To cater for partial correspondence, we filter out keypoints in the first point cloud without corresponding points in the second cloud, based on a known ground truth alignment.
$\mathcal{D'} \in \mathbb{R}^{N' \times d}$ is a filtered local descriptors matrix in the first point cloud.
$C = \mathcal{D'} \widetilde{\mathcal{D}}^{T} \in \mathbb{R}^{N' \times M}$ denotes a  correspondence matrix between local descriptors in two point clouds, where $C_{ij} \in \left[ -1, 1 \right]$ is a cosine similarity between descriptors $i$ and $j$.
%$C^{*} \in \left\{ 0, 1 \right\} ^{N' \times M}$ is a ground truth correspondence matrix computed using a known ground truth pose.
We use cross-entropy loss function defined as:
\begin{equation}
\mathcal{L}_{D} = - 
\frac{1}{N'}
\sum_{i=1}^{N'} \log 
\left( 
%\frac{\exp \left( \sum_{j=1}^{M} C_{i, j} C^{*}_{i, j}\right)}
\frac{\exp \left( C_{i, nn(i)} / \tau \right)}
{\sum_{j=1}^{M} \exp \left( C_{i, j} / \tau \right)}
\right) \, ,
\label{eq:loss_local}
\end{equation}
where $nn(i)$ is the index of the keypoint in the second point cloud corresponding to the $i$-th keypoint in the first point cloud based on a ground truth pose, and $\tau=0.02$ is a temperature hyperparameter.

%\todo[inline]{Describe what Beta is and why its needed - something like the temperature????}

\textbf{Implementation details.} All experiments are performed on a server with a single nVidia RTX 2080Ti GPU, 12 core AMD Ryzen Threadripper 1920X processor, 64 GB of RAM, and SSD hard drive. We use PyTorch 1.9~\cite{NEURIPS2019_9015} deep learning framework, MinkowskiEngine 0.5.4~\cite{choy20194d} auto-differentiation library for sparse tensors, PML Pytorch Metric Learning library 0.9.99~\cite{musgrave2020metric}, and efficient RANSAC and ICP implementations from Open3D 0.13.0~\cite{zhou2018open3d} library.

\section{Experimental Results}

\subsection{Datasets and Evaluation Methodology}

Our model is trained and evaluated using disjoint subsets of two large-scale datasets MulRan~\cite{gskim-2020-mulran} and Apollo-SouthBay~\cite{L3NET_2019_CVPR}. 
Moreover, we test generalization abilities on the KITTI odometry dataset \cite{Geiger2012CVPR}.

%\todo[inline]{Describe which elements are structurally similar and which are structurally dissimilar for training and evaluation.}

%\todo[inline]{Describe datasets: MulRan, Apollo-SouthBay, Kitti. Describe train/test split. Maybe add maps visualizing datasets, and size of each split.}

\textbf{MulRan dataset}~\cite{gskim-2020-mulran}
is gathered by a vehicle traveling through different trajectories in South Korea. Point clouds are acquired using Ouster OS1-64 rotating LiDAR with 120\,m range and contain about 60k points.
Each trajectory is traversed multiple times, at different times of day or year, to allow realistic evaluation of place recognition methods.
To speed up the processing, we remove uninformative ground plane points with z coordinate below $-0.9$\,m.
The dataset contains timestamped 6DoF ground-truth poses for each traversal estimated using VRS-GPS/INS integrated navigation system and a graph SLAM. %If a loop is detected, the relative poses are calculated with ICP algorithm using a local point cloud and the poses are used as a constraint in the graph SLAM.
The longest and most diverse trajectory, Sejong, is split into disjoint training and evaluation parts. 

\textbf{Apollo-SouthBay dataset}~\cite{L3NET_2019_CVPR} is collected during multiple traversals through six different routes in the southern San Francisco Bay Area. The dataset covers various environments, such as residential areas, urban downtown areas, and highways.  
The data is acquired using Velodyne HDL-64E rotating LiDAR with a similar range and number of scanned 3D points as in the MulRan dataset.
We remove points on the ground plane level with z coordinate below $-1.6$\,m.
Ground truth 6DoF poses are acquired by postprocessing readings from a high-end GNSS RTK/INS integrated navigation system.
The longest and most diverse trajectory (SunnyvaleBigLoop) is used for evaluation.
The other five shorter trajectories are included in the training split.

\textbf{Kitti odometry dataset}~\cite{Geiger2012CVPR} is acquired using a vehicle driving around the mid-size city of Karlsruhe, in rural areas, and on highways. It contains 11 sequences
with LiDAR point clouds and ground truth poses. 
The data is acquired with the same Velodyne HDL-64E LiDAR as used in the Apollo-SouthBay dataset.
$1.5$\,m threshold on the z-coordinate is used to remove ground plane points.
Ground truth poses are acquired using a high-end GNSS RTK/INS integrated navigation system.
We evaluate our approach on a sequence 00 as it contains the highest number of loops. We  build a map from the first 170 seconds of the sequence and leave the rest for queries. 

%  acquired with Velodyne HDL-64E LiDAR captured by driving around the mid-size city of Karlsruhe, in rural areas and on highways. 

To avoid processing multiple scans of the same place when the vehicle doesn't move, we ignore consecutive readings with less than 20 cm displacement.
Details of the training and evaluation splits are given in Tab.~\ref{mw:tab:splits}. In all three datasets, we observed that ground truth 6DoF poses between different traversals through the same trajectory are slightly misaligned. 
Therefore, we use an ICP~\cite{besl1992method} to refine 6DoF poses between pairs of point clouds from different traversals.

\begin{table}[htb!]
\caption{Details of training and evaluation sets.}
\begin{center}
%@{\enspace}
\begin{tabular}{@{\enspace}l@{\enspace}|@{\enspace}c@{\enspace}r@{\enspace}|@{\enspace}c@{\enspace}}
%\hline
& \begin{tabular}{@{}c@{}}Split \end{tabular}
& \begin{tabular}{@{}c@{}}Length \end{tabular}
& \begin{tabular}{@{}c@{}}Number of scans\end{tabular}
\\[2pt]
\hline
\Tstrut
MulRan: Sejong & train & 19 km & $35.9$k \\
SouthBay: excl. Sunnyvale & train &  37 km & $72.7$k \\
\hline
\Tstrut
MulRan: Sejong & test &  4 km & $3.8$k\,/\,$3.5$k  (map\,/\,query) \\
SouthBay: Sunnyvale & test &  38 km & $49.0$k\,/\,$17$k (map\,/\,query) \\
Kitti: Sequence 00 & test & 4 km & $1.6$k\,/\,$0.6$k (map\,/\,query)
\end{tabular}
\end{center}
\label{mw:tab:splits}
\end{table}

\subsubsection{Evaluation Metrics.} 
To evaluate the performance of the  global descriptor, we follow a similar evaluation protocol as in~\cite{uy2018pointnetvlad,du2020dh3d}.
Each evaluation set is split into two parts: a query and a database set, covering the same geographic area.
A query is formed from point clouds acquired during one traversal, and the database is built from data  gathered on a different day.
For each query point cloud, we find in the database candidate point clouds with the closest, in Euclidean distance sense, global descriptors.
Localization is successful if at least one of the top $N$ candidates is within $d$ meters
threshold from the query ground truth position. 
\emph{Recall@N} is defined as the percentage of correctly localized queries. 
We report Recall@1 (R@1) and Recall@5 (R@5) metrics for $d = 5$ and $20$\,m thresholds.

We use a simple two-step localization procedure to evaluate the performance of 6DoF pose estimation. 
First, for each query point cloud, we perform the coarse localization by searching the database for a point cloud with the closest global descriptor.
Since this work focuses on efficient keypoint and descriptor extraction, we do not consider more sophisticated approaches, where more database candidates are examined.
If the coarse localization is successful, that is if the returned point cloud is within 20\,m threshold from the ground truth query position, we estimate 6DoF relative pose between the query and the database point cloud.
If the coarse localization fails, we exclude the query point cloud from further evaluation.
From the query and database point cloud, we select 128 keypoints with the lowest saliency uncertainty ($\Sigma$).
We match selected keypoints in both clouds using their local descriptors and estimate their relative pose with RANSAC~\cite{fischler1981random}.
We report the 6DoF pose estimation success if an estimated pose is within 2\,m and $5^{\circ}$ threshold from the ground truth.
We calculate an average relative rotation error (RRE) and  relative translation error (RTE) for successful pose estimation events.

%\todo[inline]{How to we calculate yaw}
%We report the yaw estimation success if the estimated yaw is within $5^{\circ}5$ threshold from the ground truth.

\subsection{Results and Discussion}
\label{sec:results}

\textbf{Comparison with state of the art.}
We compare our method with handcrafted M2DP~\cite{he2016m2dp}, ScanContext~\cite{kim2018scan} and learned MinkLoc3D~\cite{Komorowski_2021_WACV}, Disco~\cite{xu2021disco} global descriptors;
FCGF~\cite{choy2019fully} and D3Feat~\cite{bai2020d3feat} local descriptors and combined local and global DH3D~\cite{du2020dh3d} descriptor.
We re-trained MinkLoc3D, DiSCO, and DH3D using a publicly released code and the same training sets as our method. For FCGF and D3Feat we use publicly available models trained on Kitti.
Except for M2DP, the same preprocessing is used, that is points on and below the ground plane level are removed based on the z coordinate.

%Experimental results on three evaluation datasets: MulRan~\cite{gskim-2020-mulran}, Apollo-SouthBay~\cite{L3NET_2019_CVPR} and Kitti~\cite{Geiger2012CVPR} are shown in Table~\ref{jk:tab:results1}.

Table~\ref{jk:tab:results1} shows coarse-level place recognition results using a global descriptor.
Our EgoNN method outperforms other descriptors on all three datasets. 
The highest advantage is for the MulRan dataset (Recall@1 with 5\,m threshold is almost 4\,pp above runner-up DiSCO). On other evaluation sets our method has a smaller but consistent advantage over the runner-up.

Table~\ref{jk:tab:results2} shows 6DoF pose estimation results. For FCGF and D3Feat local descriptors, we first use a global descriptor computed by EgoNN to find the closest point cloud in the database. 
Our method has a high success ratio (above 99.6\% for MulRan and Kitti and above 97\% for SouthBay).
For these successful cases, the mean relative translation error (RTE) is between 12-19\,cm and the rotation error (RRE) is between 0.3-0.4$^\circ$.
Our method outperforms the other integrated local and global descriptor, DH3D, by a large margin.
Poor DH3D performance can be explained by the fact that it was designed to process smaller point clouds.
Specialized local descriptors, FCGF and D3Feat, achieve slightly lower pose estimation errors, but at the expense of larger running times (see Runtime analysis section).

\begin{table*}[htb!]
\caption{Results of a coarse localization using a global descriptor. Recall@1 and Recall@5 are reported with 5 and 20\,m thresholds.}
\begin{center}
\begin{tabular}{l@{\quad}|c@{\quad}c@{\quad}c@{\quad}c@{\quad}|c@{\quad}c@{\quad}c@{\quad}c@{\quad}|c@{\quad}c@{\quad}c@{\quad}c}
%\hline
&  \multicolumn{4}{c}{MulRan} & \multicolumn{4}{c}{Apollo-SouthBay} & \multicolumn{4}{c}{Kitti} \\
&  \multicolumn{2}{c}{5\,m thresh.} & \multicolumn{2}{c}{20\,m thresh.} 
&  \multicolumn{2}{c}{5\,m thresh.} & \multicolumn{2}{c}{20\,m thresh.} 
&  \multicolumn{2}{c}{5\,m thresh.} & \multicolumn{2}{c}{20\,m thresh.} 
\\
& \begin{tabular}{@{}c@{}}R@1\end{tabular}
& \begin{tabular}{@{}c@{}}R@5\end{tabular}
& \begin{tabular}{@{}c@{}}R@1\end{tabular}
& \begin{tabular}{@{}c@{}}R@5\end{tabular}
& \begin{tabular}{@{}c@{}}R@1\end{tabular}
& \begin{tabular}{@{}c@{}}R@5\end{tabular}
& \begin{tabular}{@{}c@{}}R@1\end{tabular}
& \begin{tabular}{@{}c@{}}R@5\end{tabular}
& \begin{tabular}{@{}c@{}}R@1\end{tabular}
& \begin{tabular}{@{}c@{}}R@5\end{tabular}
& \begin{tabular}{@{}c@{}}R@1\end{tabular}
& \begin{tabular}{@{}c@{}}R@5\end{tabular}
% Ew. dodać liczbę iteracji RANSACa jak w DH3D
\\[2pt]
\hline
\Tstrut
% All learned models trained using mulran_train_tuples_Sejong01_Sejong02_2_10.pickle
% M2DP code (Matlab) is available here: https://github.com/LiHeUA/M2DP
M2DP~\cite{he2016m2dp} & 0.441 & 0.556 & 0.595 & 0.688 &  - & - & - & - &  0.945 & 0.953 & 0.945 & 0.953  \\
ScanContext~\cite{kim2018scan}  & 0.861 & 0.896 & 0.885 & 0.915 & 0.910 & 0.910 & 0.921 & 0.924 & \underline{0.963} & \underline{0.973} & 0.963 &  0.964 \\ 
% MinkLoc3D trained on Mulran using Cartesian coordinates
% --model_config ../models/minkloc3d.txt --weights ../weights/model_MinkLoc3D_20210927_0923_final.pth
MinkLoc3D~\cite{Komorowski_2021_WACV} & 0.823 & 0.944 & 0.921 & 0.966 & 0.772 & 0.938 & 0.950 & \underline{0.983} & 0.957 & 0.969 & \underline{0.977} & \underline{0.980} \\
% Pre-trained model occ_model.ckpt (trained on NCLT dataset) provided by authors
%DiSCO-PT~\cite{xu2021disco}& 0.805 & 0.897 & 0.858 & 0.913 & 0.727 & 1.6 & 1.5 &  & \\
%DiSCO-PT~\cite{xu2021disco} & 0.805 & 0.897 & 0.858 & 0.913 & 0.828 & 1.6 & 1.5 &  & \\
% DiSCO najlepsze wyniki po epoce 8 - model_disco_epoch_8.pth
%DiSCO~\cite{xu2021disco} & 0.875 & 0.950 & 0.940 & 0.964 & 0.932 & 1.5 & 1.5 &  \\
% New results (2021-10-15 trained on MulRan)
DiSCO~\cite{xu2021disco} & \underline{0.940} & \underline{0.975} & \underline{0.958} & \underline{0.981} & 
\underline{0.951} & \underline{0.966} & 0.954 & 0.970 & 0.907 & 0.913 & 0.923 & 0.945 \\
%\hline
%\Tstrut
%DH3D~\cite{du2020dh3d} & 0.324 & 0.562 & 0.589 & 0.755 & 0.645(0.056) & 0.8 (0.6) & 0.9 (0.7)& 0.320 & 41 & 33 & 1.6 & 1.1 \\
DH3D~\cite{du2020dh3d} & 0.324 & 0.562 & 0.589 & 0.755 & 0.253 & 0.496 & 0.504 & 0.707 &  0.755 & 0.911 & 0.868 & 0.961 \\
% model: model_egonn_20210916_1104_final.pth
% Mulran: test_Sejong01_Sejong02.pickle
% SouthBay: Evaluation set: test_SunnyvaleBigloop_1.0_5.pickle
% Kitti: kitti_00_eval.pickle
EgoNN (ours) & \textbf{0.983} &  \textbf{0.999} &  \textbf{0.996} &  \textbf{0.999} & 
\textbf{0.957} & \textbf{0.977} & \textbf{0.963} & \textbf{0.982} & 
\textbf{0.974} &  \textbf{0.982} &  \textbf{0.979} &  \textbf{0.987} 
\\[2pt]
\end{tabular}
\end{center}
\label{jk:tab:results1}
\end{table*}

\begin{table*}[hhtb!]
\caption{Results of a 6DoF pose estimation using local features within 2\,m and $5^\circ$ threshold. RTE=relative translation error, RRE=relative rotation error.}
\begin{center}
\begin{tabular}{l@{\quad}|c@{\quad}c@{\quad}c@{\quad}|c@{\quad}c@{\quad}c@{\quad}|c@{\quad}c@{\quad}c}
%\hline
&  \multicolumn{3}{c}{MulRan} & \multicolumn{3}{c}{Apollo-SouthBay} & \multicolumn{3}{c}{Kitti} 
\\
& \begin{tabular}{@{}c@{}}Success (\%) \end{tabular}
& \begin{tabular}{@{}c@{}}RTE (cm)\end{tabular}
& \begin{tabular}{@{}c@{}}RRE ($^\circ$)\end{tabular}
& \begin{tabular}{@{}c@{}}Success (\%) \end{tabular}
& \begin{tabular}{@{}c@{}}RTE (cm)\end{tabular}
& \begin{tabular}{@{}c@{}}RRE ($^\circ$)\end{tabular}
& \begin{tabular}{@{}c@{}}Success (\%) \end{tabular}
& \begin{tabular}{@{}c@{}}RTE (cm)\end{tabular}
& \begin{tabular}{@{}c@{}}RRE ($^\circ$)\end{tabular}
\\[2pt]
\hline
\Tstrut
EgoNN+FCGF~\cite{choy2019fully} & 0.965 & \underline{18} & \textbf{0.4} & 0.943 & \underline{13} & \underline{0.4} & 0.966 & \underline{10} & \underline{0.3} \\
EgoNN+D3Feat~\cite{bai2020d3feat} & \underline{0.991} & \textbf{16} & \textbf{0.4} & \textbf{0.975} & \textbf{12} & \textbf{0.3} & \textbf{0.998} & \textbf{7} & \textbf{0.2} \\
% Przeliczone dla DH3D dla threshold 1m 
DH3D~\cite{du2020dh3d} & 0.563 & 41 & 1.3 & 0.614 & 55 & 0.9 &  0.932 & 26 & 0.6  \\
% model: model_egonn_20210916_1104_final.pth
% Mulran: test_Sejong01_Sejong02.pickle
% SouthBay: Evaluation set: test_SunnyvaleBigloop_1.0_5.pickle
% kitti_00_eval.pickle
EgoNN (ours) & \textbf{0.996} & 19 & \textbf{0.4} & 
\underline{0.970} & 15 & \underline{0.4} &
\underline{0.997} &  12 & \underline{0.3} 
\\[2pt]
\end{tabular}
\end{center}
\label{jk:tab:results2}
\end{table*}

%Decrease in Recall@1 in Tab.~\ref{jk:tab:results1} because only query locations having a true match within each distance thresholds are considered. For 5 meter threshold 89\% of query elements in considered and for 20m. threshold all query elements are considered.
%These additional query elements are likely difficult examples which cause mean recall decrease.

\begin{figure}[tbh]
\centering
\includegraphics[width=1.\linewidth]{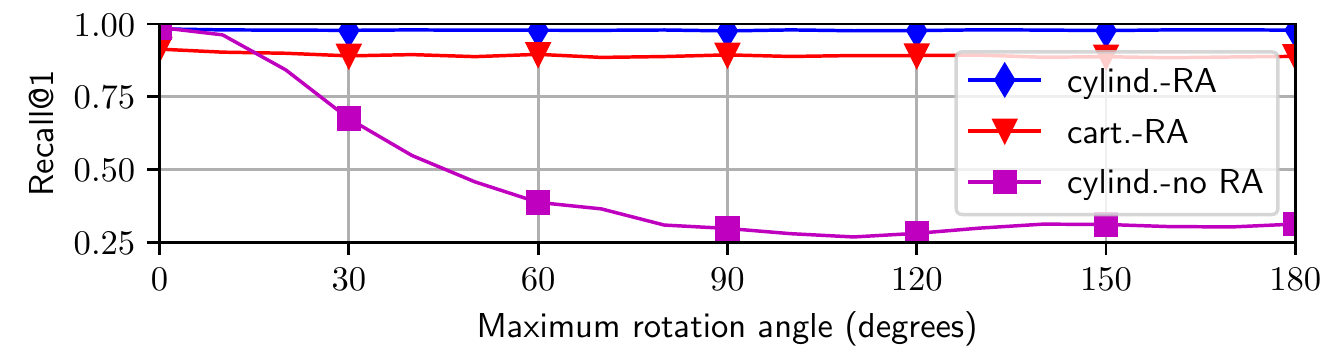}
\caption{Impact of a coordinate system choice (cylindrical vs. Cartesian) and training data augmentation (RA = random rotation) on the global descriptor rotational invariance. Plots show Recall@1 with 5\,m threshold for different angles of the orientation difference between two point clouds.
}
\label{jk:fig:rotation_impact}
\end{figure}

\textbf{Robustness to viewpoint changes.}
Figure~\ref{jk:fig:rotation_impact} shows the robustness of our global descriptor to viewpoint changes.
Plots show Recall@1 with 5\,m threshold (y-axis) on MulRan evaluation set for different magnitudes of a random rotation (x-axis) of a query point cloud.
A blue line corresponds to our EgoNN method using cylindrical coordinates and random rotation augmentation during the training. The global descriptor is invariant to the point cloud rotation.
Using Cartesian coordinates yields over 8\,pp worse Recall@1 even without applying any rotation. 
Due to the translational invariance of convolutional filters, a small point cloud translation has less effect on the global descriptor when using Cartesian coordinates instead of cylindrical. 
Thus, the nearest neighbor search in the descriptor space more likely returns more distant candidates. This deteriorates Recall@1 metric with a small 5\,m threshold.
%Thus . Thus, when finding point clouds with with closest global descriptors there's a higher change to return more distant point clouds, at a distance above 5\,m threshold.
%When cylindrical coordinates are used, global descriptor is more affected by small point cloud translations.
%So, when doing nearest neighbour search, there's a higher chance to return closest
%a global descriptor is less affected by a small point cloud translation when Cartesian coordinates are used as convolutions 
%Using Cartesian coordinates yields over 15\,pp worse results even without rotation and the gap increases to up to 20\,pp when larger rotation is applied.
Using only cylindrical coordinates, without random rotation augmentation, produces a descriptor highly sensitive to the point cloud orientation. 
%The Recall@1 is as low as 35\% for rotation angle above 90$^{circ}$.
Without random rotation augmentation, rotations still affects convolutional filter responses at the borders (in regions with $\theta$ coordinate near $0$ or $2\pi$). 
%Also, point cloud downsampling using stride 2 convolutions introduces aliasing artifacts.
%Hence, random rotation data augmentation is essential to achieve rotational invariance.

\begin{figure}[tbh]
\centering
\includegraphics[width=.85\linewidth]{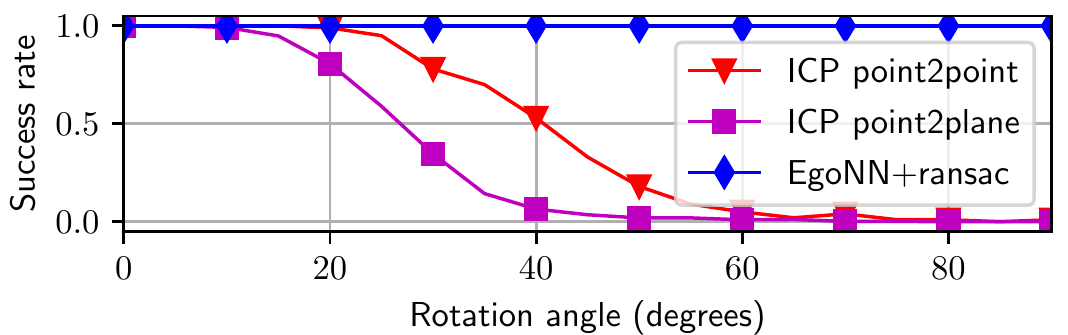}
\caption{Impact of an initial point cloud alignment on the registration success ratio. Plots show 6DoF pose estimation success ratio (with 2\,m and $5^{\circ}$ threshold) for different angles of the query point cloud rotation.
}
\label{jk:fig:rotation_impact_local}
\end{figure}

\begin{figure*}[tbh]
\centering
 \includegraphics[width=0.3\textwidth,  trim={3cm 4cm 4cm 10cm}, clip]{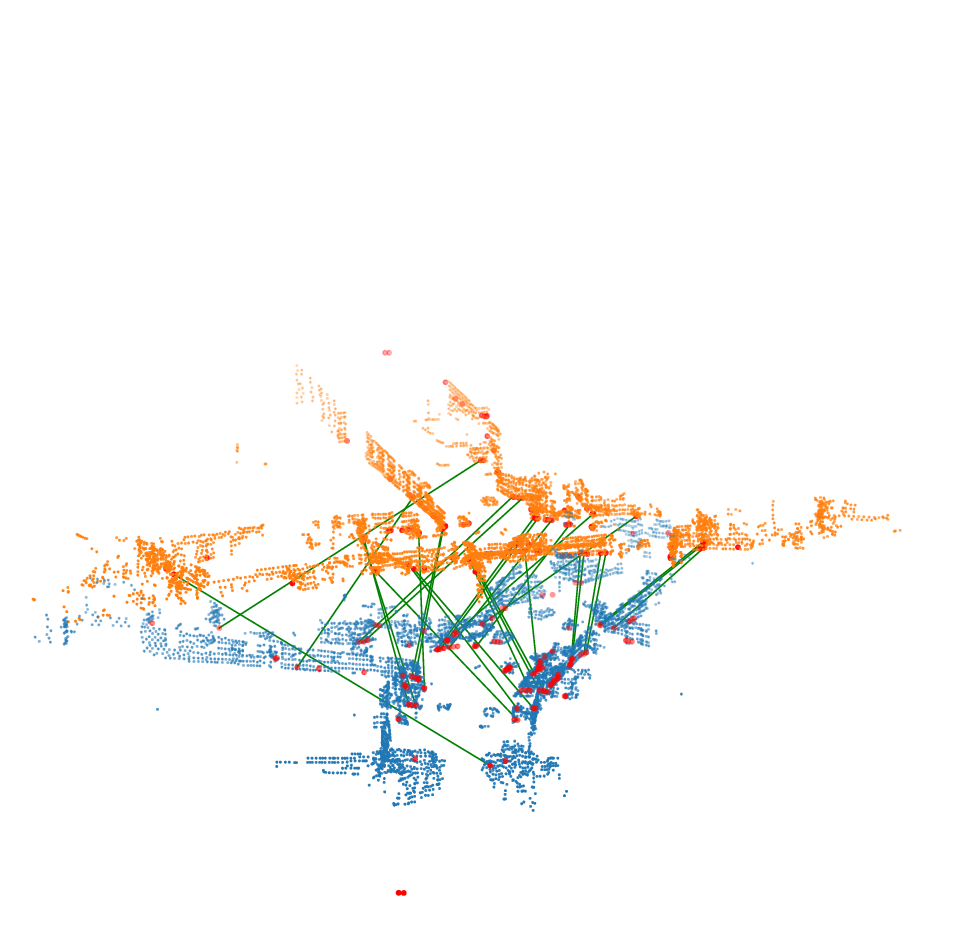}
 \hfill
 \includegraphics[width=0.3\textwidth,  trim={4cm 4cm 3cm 9cm}, clip]{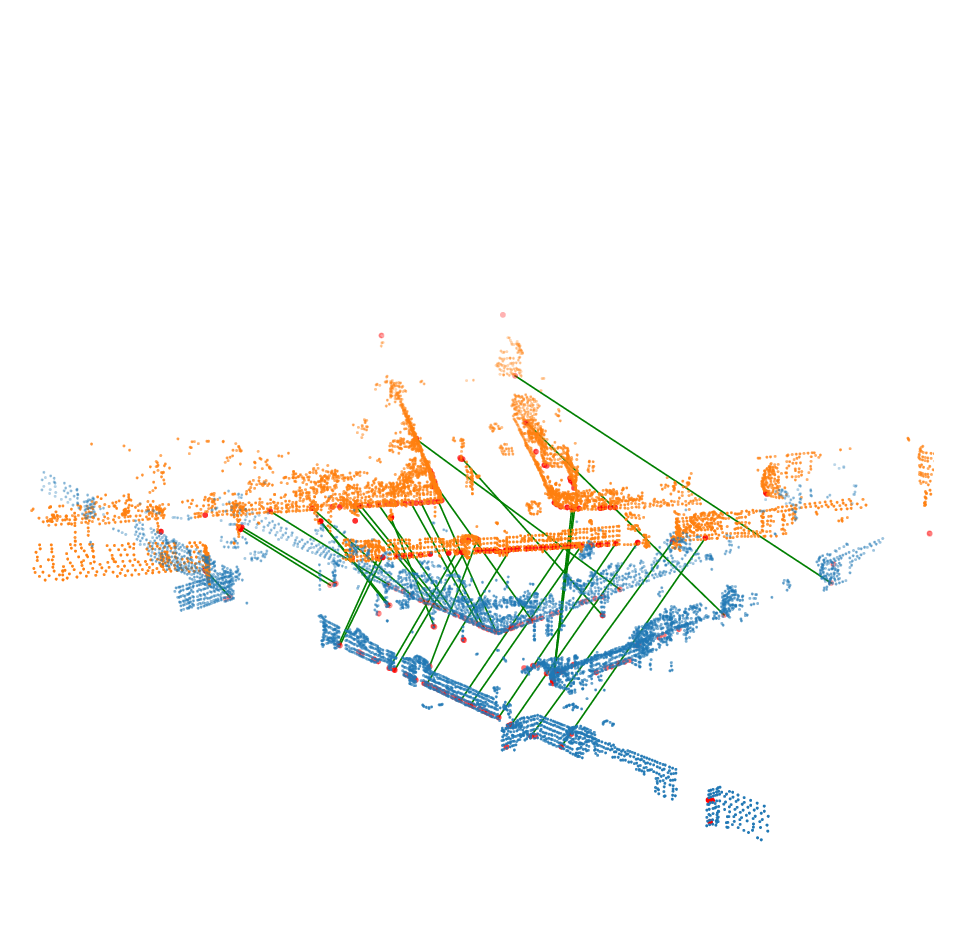}
 \hfill
 \includegraphics[width=0.3\textwidth,  trim={2cm 4cm 4cm 10cm}, clip]{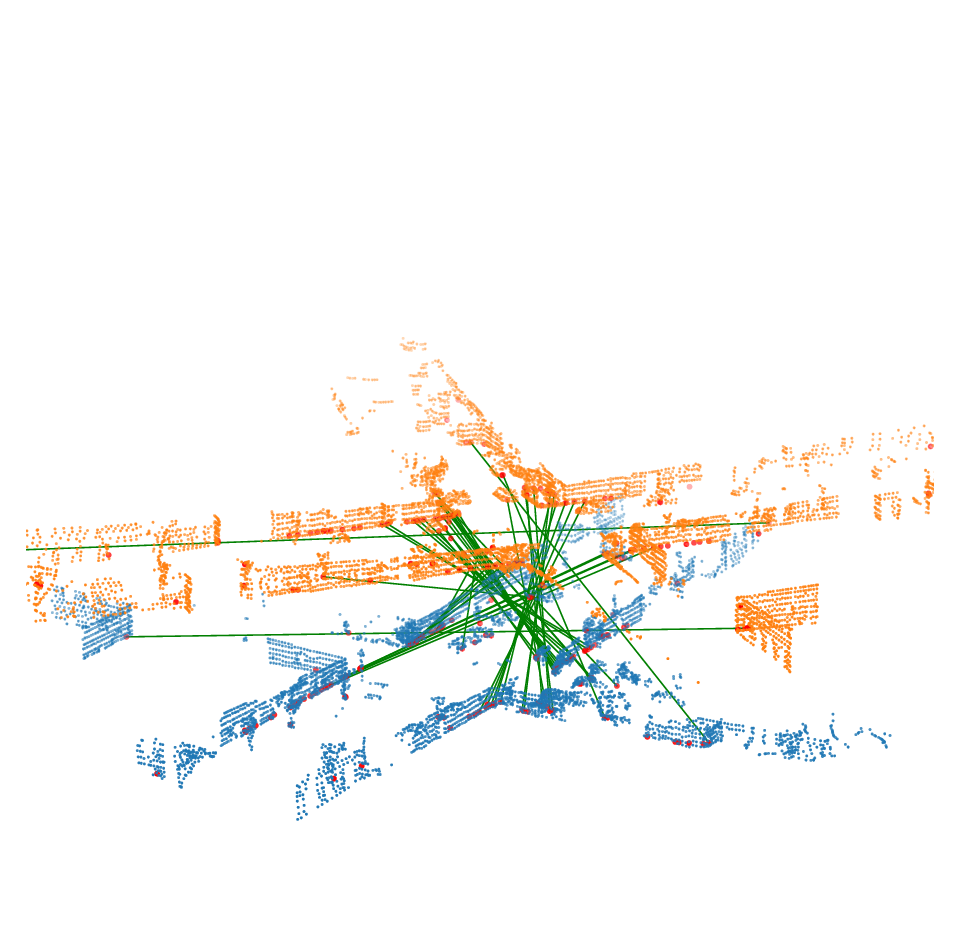}
\caption{Successful registration of the exemplary point clouds gathered during revisiting the same place from different directions in the Kitti dataset using EgoNN with RANSAC. Green lines connect corresponding keypoints (RANSAC inliers).}
\label{jk:fig:registration}
\end{figure*}

Figure~\ref{jk:fig:rotation_impact_local} shows 6DoF pose estimation success ratio within 2\,m and $5^{\circ}$ threshold as a function of an initial orientation of two point clouds.
Both point-to-point and point-to-plane ICP~\cite{besl1992method} variants are sensitive to an initial point cloud alignment. Success ratio starts dropping down when initial orientation difference is larger than $20^{\circ}$.
On the other hand, using EgoNN keypoints and descriptors with RANSAC allows successful 6DoF pose estimation despite a large initial misalignment.
Examples are visualized in Fig.~\ref{jk:fig:registration}. 
It show point cloud pairs gathered during revisiting same places from different directions in the Kitti dataset. 
In all three cases, 6DoF pose estimation error was below 16\,cm and $0.8^{\circ}$, whereas ICP registration failed and got stuck in a local minima due to a large initial misalignment.

\begin{figure}[tbh]
\centering
%\subfloat{%
%\includegraphics[width=8.83cm,height=4.6cm,trim={1cm 0cm 0cm 0cm},clip,cfbox=black 0.5pt 0.5pt]{images/key1_.png}
%\includegraphics[width=8.83cm,height=4.6cm,trim={1cm 0cm 0cm 0cm},clip]{images/key1_.png}
\includegraphics[width=0.85\linewidth,trim={1cm 1.3cm 0cm 0cm},clip]{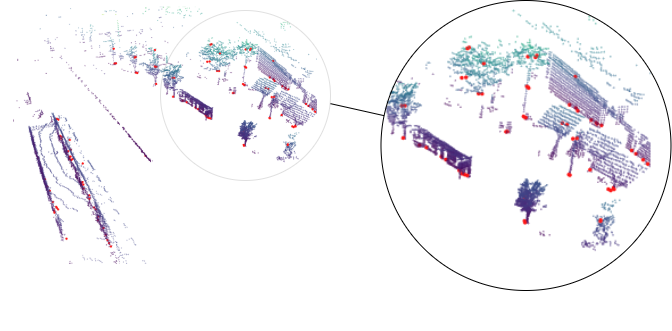}
%}
%\hfill
%\subfloat{%
%\includegraphics[width=8.83cm,height=4.6cm,trim={0cm 0cm 0cm 1cm},clip,cfbox=black 0.5pt 0.5pt]{images/key2_.png}
%\includegraphics[width=8.83cm,height=4.6cm,trim={0cm 0cm 0cm 1cm},clip]{images/key2_.png}
%}
\caption{Visualization of keypoint detection results, showing 128 points with the lowest saliency uncertainty.}
\label{fig:keypoints_vis}
\end{figure}

\begin{figure}[tbh]
\centering
\includegraphics[width=1.0\linewidth]{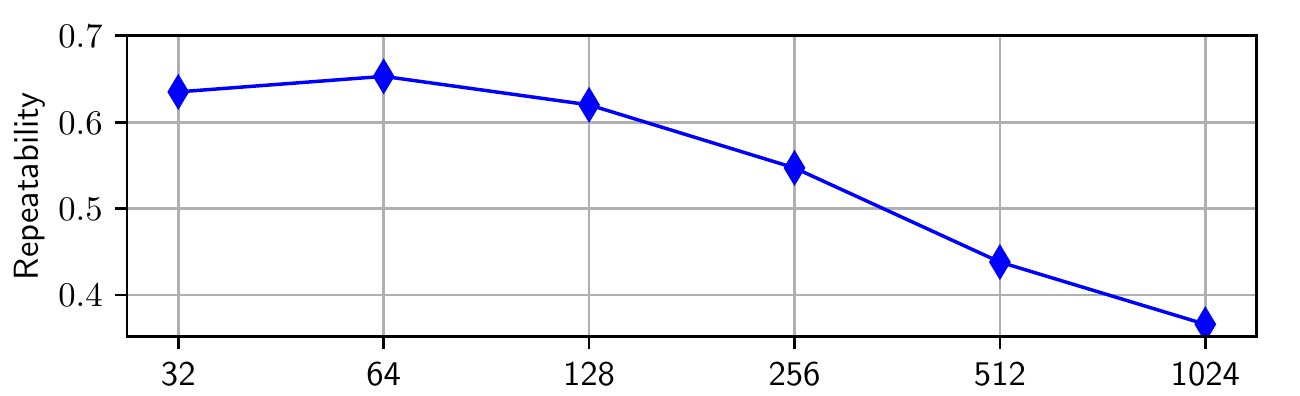}
\caption{Repeatability when a different number of keypoints with the lowest saliency uncertainty is chosen.}
\label{fig:repeatability}
\end{figure}

\textbf{Keypoint detection.} Figure~\ref{fig:keypoints_vis} shows 128 keypoints with the lowest saliency uncertainty $\Sigma$ detected in an exemplary point cloud. 
It can be seen that keypoints are detected at distinctive locations, such as the bottom of tree trunks.
The repeatability of keypoint detector is examined in Figure~\ref{fig:repeatability}. A keypoint is repeatable if, in the other point cloud aligned using the ground truth transform, another keypoint is detected within 0.5\,m threshold. 
Our method generates highly repeatable keypoints. When 64 keypoints with the lowest saliency uncertainty $\Sigma$ are selected, repeatability is as high as 65\% and decreases to 37\% for 1024 most salient keypoints.

\textbf{Ablation study} results are shown in Table~\ref{jk:tab:ablation}.
The top row shows the performance of our method on the MulRan evaluation set.
Lower rows show results with parts of the network removed or disabled.
Simplifying the network architecture by removing the top-down processing path (no top-down blocks), containing blocks 
{\tt Conv$_6$, Conv$_7$ TConv$_6$, TConv$_7$, TConv$_4$}, lowers the coarse localization performance by 3-5\,p.p. 
Disabling keypoint saliency uncertainty estimation (no saliency regress.), and selecting 128 random keypoints instead of keypoints with the lowest uncertainty, almost doubles relative translation error and increases rotation error by 50\%.
Removing the keypoint position regressor (no position regress.), and assuming keypoint locations at the supervoxel centers, further degrades the performance of 6DoF pose estimation.

%Below results are for evaluation results on Mulran dataset (map: Sejong01, query: Sejong02)
\begin{table}[hhtb!]
\caption{Ablation study. The top row is a full EgoNN model. Lower rows show the performance of reduced architectures.}
\begin{center}
\begin{tabular}{l@{\enskip}|@{\enskip}c@{\enskip}c|@{\enskip}c@{\enskip}c@{\enskip}c}
%\hline
&  \multicolumn{2}{c}{COARSE LOC.} & \multicolumn{3}{c}{6DoF POSE EST.} \\
& 5 m
&  20 m
& \multicolumn{3}{c}{2\,m and $5^\circ$ threshold} \\
& R@1
&  R@1
& \begin{tabular}{@{}c@{}}Succ.(\%) \end{tabular}
& \begin{tabular}{@{}c@{}}RTE(cm)\end{tabular}
& \begin{tabular}{@{}c@{}}RRE($^\circ$)\end{tabular}
\\[2pt]
\hline
\Tstrut
% model_egonn_20210916_1104_final.pth
EgoNN & \textbf{0.983} & \textbf{0.996} & \textbf{0.996}  & \textbf{19} & \textbf{0.4} \\
% model_egonn_nofpn1_20210918_0058_final.pth
no top-down blocks & 0.934 & 0.964 & 0.989 & \textbf{19} & \textbf{0.4} \\
no saliency regress. & \textbf{0.983} & \textbf{0.996} & 0.988 & 34 & 0.6 \\
no position regress. & \textbf{0.983} & \textbf{0.996} & 0.854 & 67 & 0.8 \\
%\\[2pt]
%\Tstrut
%model_minkgl3_505_20210616_1436 
%EgoNN-cart. & 0.838 & 0.976 &  & 22 & 0.4 \\
%\\[2pt]
\end{tabular}
\end{center}
\label{jk:tab:ablation}
\end{table}

%Compare results for different dimensionality of global and local descriptors.
% For kitti point cloud with torch.Size([70241, 3]) quantized to torch.Size([13187, 4])
% Total time: 28.168113231658936   per iter: 28.2 [ms]

\textbf{Runtime analysis.} 
EgoNN requires 28\,ms for extraction of global descriptor and local keypoints from a point cloud with 30k points. Other method extracting both global and local descriptors, DH3D~\cite{du2020dh3d}, requires 78\,ms.
%It must be noted that DH3D operates on clouds downsampled to 8192 points, whereas our method operates on much larger scans from a 3D rotating LiDAR.
Local descriptor extraction methods require much longer time: 
360\, ms for FCGF~\cite{choy2019fully} and 130\,ms for D3Feat~\cite{bai2020d3feat}.
Methods extracting only a global descriptor are faster: 8\,ms for ScanContext, 10\,ms for DiSCO, 14\,ms for MinkLoc3D and 22\,ms for M2DP. 
But 10-20\,ms extra time needed by EgoNN is acceptable, %It can run at 35\,Hz.
as it additionally produces repetitive keypoints and discriminative local descriptors allowing a robust 6DoF pose estimation. 
It takes only 5\,ms to match 128 keypoints with the lowest saliency uncertainty and estimate 6DoF pose between two point clouds with RANSAC.
%Total time needed for 6DoF pose estimation is 33\,ms on average: 28\,ms for keypoints and descriptors extraction plus 5\,ms for matching 128 keypoints with the lowest saliency uncertainty and running robust RANSAC estimator. 
%In contrast, registering two same size point clouds using ICP takes .... on average.

%This extra 10-20\,ms needed by our method is offset by the fact that in addition to a global descriptor, our method computes  repetitive keypoints and discriminative local descriptors. 
%This allows quick 6DoF pose estimation with RANSAC.....................................

The efficiency of EgoNN can be attributed to three factors. First, computations during a local and global descriptor extraction are shared. 
Second, we do not compute local descriptors for all 3D points, but only for much lower number of detected keypoints.
Third, using sparse voxelized representation and relatively simple convolutional architecture allows efficient processing with MinkowskiEngine~\cite{choy20194d} auto-differentiation library for sparse tensors.

\section{Conclusion}

This paper presents EgoNN, a 3D convolutional architecture based on a sparse voxelized representation, to efficiently extract global and local descriptors from large point clouds acquired using a modern 3D rotating LiDAR.
Experimental evaluation proves that our global descriptor outperforms prior cloud-based place recognition methods. Local descriptors allow efficient 6DoF pose estimation within 19\,cm and 0.4$^\circ$ degree accuracy.
%The success of our method can be attributed to the efficient design. 
%Using sparse volumetric representation allows efficient processing using 3D convolutional architecture and extract informative features that can be aggregated into a discriminative global descriptor.
For simplicity, we evaluated the performance of a coarse localization using only global descriptors. 
A potential research direction is to investigate how the localization accuracy can be improved by re-ranking candidate locations using local keypoints and descriptors.


%%%%%%%%%%%%%%%%%%%%%%%%%%%%%%%%%%%%%%%%%%%%%%%%%%%%%%%%%%%%%%%%%%%%%%%%%%%%%%%%

\bibliographystyle{IEEEtran}
\bibliography{jk-bib}

\end{document}